\documentclass[11pt,a4paper]{article}
\usepackage[hyperref]{naaclhlt2019}
\usepackage{times,latexsym}
\usepackage{url}
\usepackage[T1]{fontenc}
\usepackage{float}
\usepackage{enumitem}
\usepackage{amsmath,amsfonts,amssymb,amsthm}
\usepackage{bbm}
\usepackage{verbatim}
\usepackage{todonotes}
\usepackage{tabularx}
\usepackage{subcaption}
\usepackage{pgfplotstable}
\usepackage{booktabs}
\pgfplotsset{compat=1.7}
\usepackage{algorithm}
\usepackage{algpseudocode}

\usepackage{datatool}
\usepackage{bashful}

\usepackage{xcolor}
\usepackage{amssymb}
\usepackage{wasysym}
\usepackage{siunitx}
\usepackage{calc}
\usepackage{caption}

\aclfinalcopy % Uncomment this line for the final submission
\title{An Analysis of Attention over Clinical Notes for Predictive Tasks}

\author{Sarthak Jain \\
  Northeastern University \\
  {\small\tt jain.sar@husky.neu.edu} \\\And
  Ramin Mohammadi \\
  Northeastern university \\
  {\small\tt mohammadi.r@husky.neu.edu} \\\And
  Byron C. Wallace \\
  Northeastern University \\
  {\small\tt b.wallace@northeastern.edu} \\}

\date{}

\newcommand{\vect}[1]{\boldsymbol{\mathbf{#1}}}
\newcommand{\R}{\mathbb{R}}

\newcommand{\AddSubFigure}[4]{
\begin{subfigure}[b]{.18\linewidth}
    \includegraphics[width=\linewidth]{#1/#2+#3.pdf}
    \caption{#4}
\end{subfigure}
}
\newcommand{\trunc}[1]{\num[round-mode=places,round-precision=2]{#1}}

\newcommand{\AddSubFigureForAll}[1]{
\AddSubFigure{#1}{readmission}{lstm+tanh}{Readmission}
\AddSubFigure{#1}{mortality}{lstm+tanh}{Mortality}
\AddSubFigure{#1}{KneeSurgery}{lstm+tanh}{Knee Surgery}
\AddSubFigure{#1}{HipSurgery}{lstm+tanh}{Hip Surgery}
\AddSubFigure{#1}{Phenotyping}{lstm+tanh}{Phenotyping}
}

\newcommand{\AddSubFigureScatter}[1]{
\AddSubFigure{#1}{readmission}{lstm+tanh}{Readmission}
\AddSubFigure{#1}{mortality}{lstm+tanh}{Mortality}
\AddSubFigure{#1}{KneeSurgery}{lstm+tanh}{Knee Surgery}
\AddSubFigure{#1}{HipSurgery}{lstm+tanh}{Hip Surgery}
\AddSubFigure{#1}{Phenotyping}{lstm+tanh}{Phenotyping}

\AddSubFigure{#1}{readmission}{cnn_1,3,5,7_+tanh}{Readmission}
\AddSubFigure{#1}{mortality}{cnn_1,3,5,7_+tanh}{Mortality}
\AddSubFigure{#1}{KneeSurgery}{cnn_1,3,5,7_+tanh}{Knee Surgery}
\AddSubFigure{#1}{HipSurgery}{cnn_1,3,5,7_+tanh}{Hip Surgery}
\AddSubFigure{#1}{Phenotyping}{cnn_1,3,5,7_+tanh}{Phenotyping}

\AddSubFigure{#1}{readmission}{average+tanh}{Readmission}
\AddSubFigure{#1}{mortality}{average+tanh}{Mortality}
\AddSubFigure{#1}{KneeSurgery}{average+tanh}{Knee Surgery}
\AddSubFigure{#1}{HipSurgery}{average+tanh}{Hip Surgery}
\AddSubFigure{#1}{Phenotyping}{average+tanh}{Phenotyping}
}

\newcommand{\DTLloadgrad}[1]{
\DTLloaddb{grad_lstm_#1}{graph_outputs/GradientPval/#1+lstm+tanh.csv}
\DTLloaddb{grad_average_#1}{graph_outputs/GradientPval/#1+average+tanh.csv}
}

\newcommand{\generaterow}[3]{
\DTLforeach{grad_#3_#1}{\Class=Column1,\mean=mean,\std=std,\pval=pval_sig}%
{%
  \DTLiffirstrow{#2}{} & \Class & \trunc{\mean} $\pm$ \trunc{\std} & \trunc{\pval}  \DTLiflastrow{\hspace{-4pt}}{\\} 
}%
}

\newcommand{\GradTabs}{
\DTLloadgrad{readmission}
\DTLloadgrad{mortality}
\DTLloadgrad{KneeSurgery}
\DTLloadgrad{HipSurgery}
\DTLloadgrad{Phenotyping}
\begin{table}
\small
\centering
\begin{tabular}{cccc}
Dataset & Class & Mean $\pm$ Std. & Sig. Frac. \\ \hline
\textbf{LSTM Encoder} & & & \\
\generaterow{readmission}{Readmission}{lstm} \\
\generaterow{mortality}{Mortality}{lstm} \\
\generaterow{KneeSurgery}{Knee Surgery}{lstm} \\
 \generaterow{HipSurgery}{Hip Surgery}{lstm} \\
 \generaterow{Phenotyping}{Phenotyping}{lstm} \\ \hline
 
\textbf{Projection Encoder} & & & \\
\generaterow{readmission}{Readmission}{average} \\
\generaterow{mortality}{Mortality}{average} \\
\generaterow{KneeSurgery}{Knee Surgery}{average} \\
 \generaterow{HipSurgery}{Hip Surgery}{average} \\
 \generaterow{Phenotyping}{Phenotyping}{average}

\end{tabular}
\vspace{-.5em}
\caption{Mean and std. dev. of correlations between gradient importance measures and attention weights. \emph{Sig. Frac.} columns report the fraction of instances for which this correlation is statistically significant.} 
    \label{tab:tau-sig}
\vspace{-1em}
\end{table}
}

\newcommand{\DTLloadevals}[1]{
\DTLloaddb{eval_#1}{graph_outputs/evals/#1+lstm+tanh.csv}
}

\newcommand{\mybar}[1]{%%
  #1 {\color{black}\rule{#1cm}{8pt}}{\color{lightgray!30}\rule{1cm-#1cm}{8pt}}
  }
  
\newcommand{\generateroweval}[2]{
\textbf{#2} & & \\
\DTLforeach{eval_#1}{\Class=Model,\roc=roc_auc,\pr=pr_auc}%
{%
  \Class & \DTLround{\roc}{\roc}{2}\mybar{\roc} & \DTLround{\pr}{\pr}{2}\mybar{\pr} \DTLiflastrow{\hspace{-1pt}}{\\} 
}%
}

\newcommand{\EvalTabs}{
\DTLloadevals{readmission}
\DTLloadevals{mortality}
\DTLloadevals{KneeSurgery}
\DTLloadevals{HipSurgery}
\DTLloadevals{Phenotyping}
\begin{table}[h!]
\small
\centering
\begin{tabular}[t]{b{3.8cm}@{\hskip 0.07in}l@{\hskip 0.03in}l@{\hskip 0.0in}}
Model & ROC AUC & PR AUC \\ \hline
\generateroweval{readmission}{Readmission} \\ \hline
\generateroweval{mortality}{Mortality} \\\hline
\generateroweval{KneeSurgery}{Knee Surgery Complication} \\\hline
\generateroweval{HipSurgery}{Hip Surgery Complication} \\\hline
\generateroweval{Phenotyping}{Phenotyping}

\end{tabular}
\vspace{-.5em}
\caption{Predictive results across all datasets and tasks using different models and attention variants.} 
    \label{tab:evals}
\vspace{-.5em}
\end{table}
}

\begin{document}
\maketitle
\begin{abstract}
The shift to electronic medical records (EMRs) has engendered research into machine learning and natural language technologies to analyze patient records, and to predict from these clinical outcomes of interest. 
Two observations motivate our aims here. 
First, unstructured notes contained within EMR often contain key information, and hence should be exploited by models.
Second, while strong predictive performance is important, \emph{interpretability} of models is perhaps equally so for applications in this domain.
Together, these points suggest that neural models for EMR may benefit from incorporation of \emph{attention} over notes, which one may hope will both yield performance gains and afford transparency in predictions.
In this work we perform experiments to explore this question using two EMR corpora and four different predictive tasks, that: (i) inclusion of \emph{attention mechanisms} is critical for neural encoder modules that operate over notes fields in order to yield competitive performance, but, (ii) unfortunately, while these boost predictive performance, it is decidedly less clear whether they provide meaningful support for predictions. Code to reproduce all experiments is available at {\small \url{https://github.com/successar/AttentionExplanation}}.
\end{abstract}

\section{Introduction}

The adoption of electronic medical records (EMRs) has spurred development of machine learning (ML) and natural language processing (NLP) methods that analyze the data these records contain; for a recent survey of such efforts, see \cite{shickel2018deep}. 
Key information for downstream predictive tasks (e.g., forecasting whether a patient will need to be readmitted within 30 days) may be contained within unstructured notes fields \cite{boag2018s,jin2018improving}. 

In this work we focus on the modules within neural network architectures responsible for encoding text (notes) into a fixed-size representation for consumption by downstream layers. 
Patient histories are often long and may contain information mostly irrelevant to a given target. Encoding this may thus be difficult, and text encoder modules may benefit from \emph{attention mechanisms} \cite{bahdanau2014neural}, which may be imposed to emphasize relevant tokens. 

In addition to mitigating noise introduced by irrelevant tokens, attention mechanisms are often seen as providing interpretability, or insight into model behavior. 
However, recent work \cite{jain2019attention} has argued that treating attention as explanation may, at least in some cases, be misguided. 
Interpretability is especially important for clinical tasks, but incorrect or misleading rationales supporting predictions may be particularly harmful in this domain; this motivates our focused study in this space.

To summarize, our {\bf contributions} are as follows. First, we empirically investigate whether incorporating standard attention mechanisms into RNN-based text encoders improves the performance of predictive models learned over EMR. We find that they do; inclusion of standard additive attention mechanism in LSTMs consistently yields absolute gains of $\sim$10 points in AUC, compared to an LSTM without attention.\footnote{Indeed, across both corpora and all tasks considered, inattentive LSTMs perform considerably worse than logistic regression and bag-of-words (BoW); introducing attention makes the neural variants competitive, but not decisively better. We hope to explore this point further in future work.} Second, we evaluate the induced attention distributions with respect to their ability to `explain' model predictions. We find mixed results here, similar to \cite{jain2019attention}: attention distributions correlate only weakly (though almost always significantly) with gradient measures of feature importance, and we are often able to identify very different attention distributions that nonetheless yield equivalent predictions. Thus, one should not in general treat attention weights as meaningful explanation of predictions made using clinical notes.

\section{Models}
We experiment with multiple standard encoding architectures, including: (i) a standard BiLSTM model; (ii) a convolutional model, and (iii) an embedding projection based model. We couple each of these with an attention layer, following \cite{jain2019attention}. 
Concretely, each encoder yields hidden state vectors $\{h_1, ... , h_T\}$, and an attention distribution $\{\alpha_1, ... , \alpha_T \}$ is induced over these according to a scoring function $\phi$: $\hat{\vect{\alpha}} = \text{softmax}(\phi (\vect{h})) \in \R^{T}$. In this work we consider  \emph{Additive} similarity functions $\phi(\vect{h}) = \vect{v}^T\text{tanh}(\vect{W_1} \vect{h} + \vect{b})$ \cite{bahdanau2014neural}, where $\vect{v}, \vect{W_1}, \vect{W_2}, \vect{b}$ are model parameters. Predictions are made on the basis of induced representations: $\hat{y} = \sigma(\vect{\theta} \cdot h_\alpha) \in \R^{|\mathcal{Y}|}$, where $h_\alpha = \sum_{t=1}^T \hat{\alpha}_t \cdot h_t$ and $\vect{\theta}$ are top-level discriminative (e.g., softmax) parameters.

\section{Datasets and Tasks} 

We consider five tasks over two independent EMR datasets. The first EMR corpus is MIMIC-III \cite{johnson2016mimic}, a publicly available set of records from patients in the Intensive Care Unit (ICU). We follow prior work in modeling aims and setup on this dataset. Specifically we consider the following predictive tasks on MIMIC.

\begin{enumerate}[leftmargin=*]
    \item \textbf{Readmission}. The task here is to predict patient readmission within 30 days of discharge or transfer from the ICU. We follow the cohort selection of \cite{Lin385518}. We assume the model has access to all notes from patient admission up until the discharge or transfer from the ICU (the point of prediction).
    
    \item \textbf{Retrospective 1-yr mortality}. We aim to predict patient mortality within one year. In this we follow the experimental setup of \cite{ghassemi2014unfolding}. The model is provided all notes up until patient discharge (excluding the discharge summary). 
       
      % bcw -- can you say a bit more about what these are? like phenotype == specific icd code? or something else? 
      %sarthak: phenotypes are a subset of hcup ccs classification into which icd9 are grouped. 
    \item \textbf{Phenotyping}. Here we aim to predict the top 25 acute care phenotypes for patients (associated at discharge with the admission). For this we again rely on the framing established in prior work \cite{harutyunyan2017multitask}. The model has access to all notes from admission up until the end of the ICU stay.
    Note that this may be viewed as a multilabel classification task, similar to \cite{harutyunyan2017multitask,lipton2015learning}.
    
\end{enumerate}

% bcw 4/4: @Sarthak can we provide more details here?? what hospital? maybe also add the IRB number in a footnote here if we can get it
The second EMR dataset we use comprises records for 7174 patients from Mass General Hospital who underwent hip or knee arthroplasty procedures. Use of this data was approved by an Institutional Review Board (IRB protocol number 2016P002062) at Partners Healthcare.
%\footnote{We withhold additional details concerning this dataset to preserve anonymization for submission and review, however we will provide specifics prior to publication. Use of this data was approved by an appropriate Institutional Review Board.}  

\begin{enumerate}[leftmargin=*]
    \item \textbf{Predicting Hip and Knee Surgery Complications}. We consider patients who underwent hip or knee arthroplasty procedure; we aim to classify these patients with respect to whether or not they will be readmitted within 30 days due to surgery-related complications. We run experiments over hip and knee surgery patients separately.   
\end{enumerate}

\section{Experiments}

\begin{figure*}
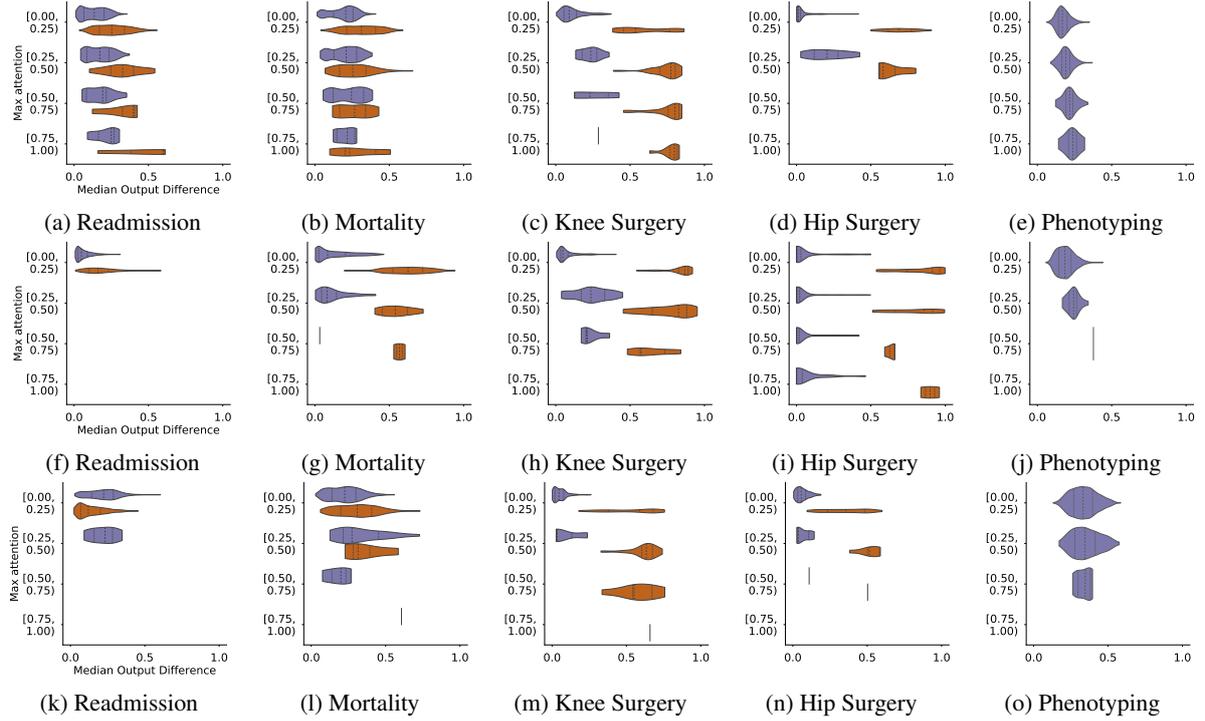
%[!hbtp]
    \centering
    \AddSubFigureScatter{graph_outputs/Permutation_MAvDY} %\vspace{-.5em}
    \caption{\textbf{Median change in output $\boldsymbol{\Delta\hat{y}^{med}}$} (x) densities in relation to the \textbf{max attention ($\boldsymbol{\max{\hat{\alpha}}}$)} (y) obtained by randomly permuting instance attention weights. Colors denote classes: negative (\textcolor[HTML]{7570b3}{$\blacksquare$}) and positive (\textcolor[HTML]{d95f02}{$\blacksquare$}); phenotyping (e) is not binary. {\bf Top row shows results for BiLSTM encoders; middle for CNNs; bottom for Embedding Projection.} }
    \label{fig:PermutationScatter}
        \vspace{-.5em}
\end{figure*}

\begin{figure*}
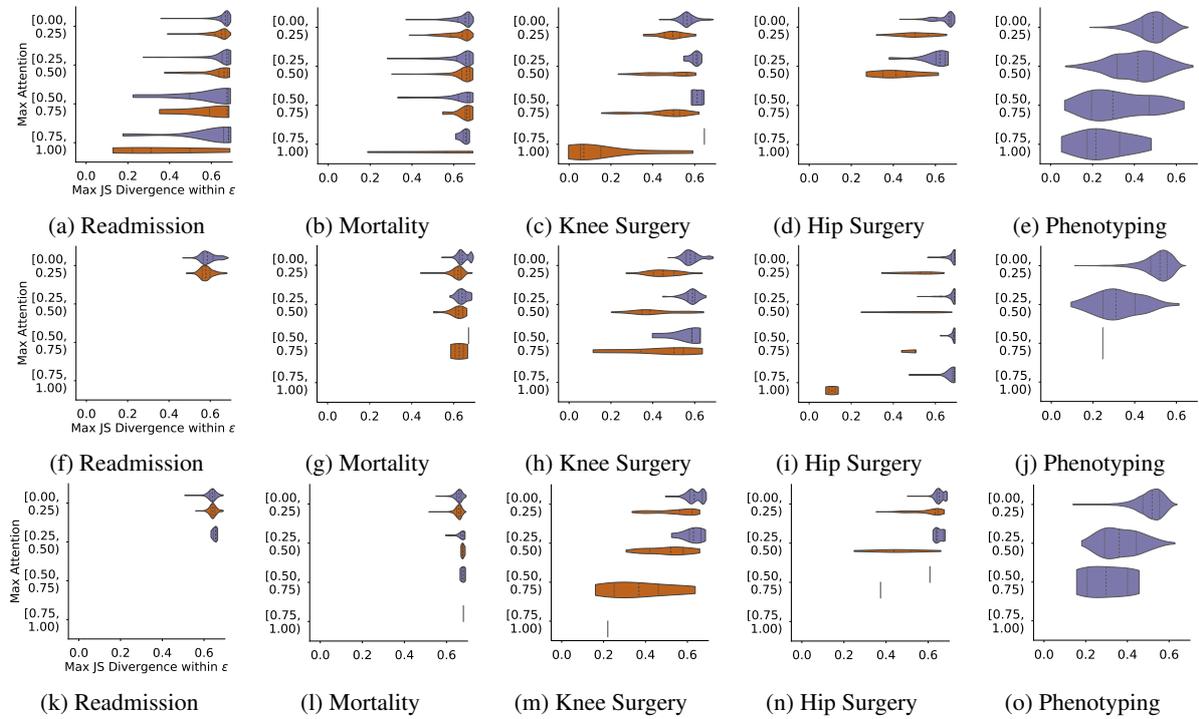
%[!hbtp]
    \centering
    \AddSubFigureScatter{graph_outputs/eMaxJDS_Scatter} \vspace{-.5em}
    \caption{Densities of \textbf{maximum JS divergences ($\epsilon\text{-max JSD}$)} (x-axis) as a function of the \textbf{max attention} (y-axis) in each instance for obtained between original and adversarial attention weights. Colors are as above. {\bf Top row shows results for BiLSTM encoders; middle for CNNs; bottom for Embedding Projection.}}
    \label{fig:AdversarialScatter}
   %\vspace{-1em}
\end{figure*}

Following the analysis of \cite{jain2019attention} but focusing on clinical tasks, we perform a set of experiments on these corpora that aim to assess the degree to which attention mechanisms aid (or hamper) predictive performance, and the degree to which the induced attention weights might be viewed as providing explanations for predictions. 

The latter can be assessed in many ways, depending on one's view of interpretability. 
To address the question of whether it is reasonable to treat attention as providing interpretability broadly, we perform experiments that interrogate multiple properties we might expect these weights to exhibit if so. Specifically, we: probe the degree to which attention weights correlate with alternative gradient-based feature importance measures, which have a more straight-forward interpretation \cite{ross2017right,li2016understanding}; evaluate whether we are able to identify `counterfactual' attention distributions that change the attention weights (focus) but not the prediction; and, in an exercise novel to the present work, we consider replacing attention weights with log odds scores from a logistic regression (linear) model. We provide a web interface to interactively browse the plots for all datasets, model variants, and experiment types: {\small\url{https://successar.github.io/AttentionExplanation/docs/}}.

 \subsection{Gradient Experiments}
To evaluate correlations between attention weights and gradient based feature importance scores, we compute Kendall-$\tau$ measure (\ref{tab:tau-sig}) between attention scores and gradients with respect to the tokens comprising documents. Across both corpora and all tasks we observe only a modest correlation between the two for BiLSTM model (the projection based model have higher correspondence, which is expected for such simple architectures). This may be problematic for attention as an explanatory mechanism, given the explicit relationship between gradients and model outputs. (Although we note that gradient based methods themselves pose difficulty with respect to interpretation \cite{feng2018pathologies}). 

\GradTabs

\begin{figure*}
\begin{minipage}{\textwidth}
\small
\textbf{Original vs Adversarial Attention Difference :} 
{\setlength{\fboxsep}{0pt}\colorbox[Hsb]{350, 0.00, 1.0}{\strut Sed}} 
{\setlength{\fboxsep}{0pt}\colorbox[Hsb]{350, 0.00, 1.0}{\strut dolorem}} 
{\setlength{\fboxsep}{0pt}\colorbox[Hsb]{350, 0.00, 1.0}{\strut sed}} 
{\setlength{\fboxsep}{0pt}\colorbox[Hsb]{202, 0.00, 1.0}{\strut adipisci}} 
{\setlength{\fboxsep}{0pt}\colorbox[Hsb]{202, 0.00, 1.0}{\strut ipsum}} {\setlength{\fboxsep}{0pt}\colorbox[Hsb]{350, 0.00, 1.0}{\strut dolor}} {\setlength{\fboxsep}{0pt}\colorbox[Hsb]{350, 0.00, 1.0}{\strut dolorem.}} {\setlength{\fboxsep}{0pt}\colorbox[Hsb]{350, 0.00, 1.0}{\strut Ut}} {\setlength{\fboxsep}{0pt}\colorbox[Hsb]{202, 0.00, 1.0}{\strut adipisci}} {\setlength{\fboxsep}{0pt}\colorbox[Hsb]{350, 0.00, 1.0}{\strut magnam}} {\setlength{\fboxsep}{0pt}\colorbox[Hsb]{350, 0.00, 1.0}{\strut tempora.}} {\setlength{\fboxsep}{0pt}\colorbox[Hsb]{350, 0.00, 1.0}{\strut Modi}} {\setlength{\fboxsep}{0pt}\colorbox[Hsb]{350, 0.01, 1.0}{\strut \#}} {\setlength{\fboxsep}{0pt}\colorbox[Hsb]{202, 0.00, 1.0}{\strut eius}} {\setlength{\fboxsep}{0pt}\colorbox[Hsb]{350, 0.07, 1.0}{\strut :}} {\setlength{\fboxsep}{0pt}\colorbox[Hsb]{350, 0.00, 1.0}{\strut tempora}} {\setlength{\fboxsep}{0pt}\colorbox[Hsb]{350, 0.03, 1.0}{\strut change}} {\setlength{\fboxsep}{0pt}\colorbox[Hsb]{350, 0.00, 1.0}{\strut ipsum}} {\setlength{\fboxsep}{0pt}\colorbox[Hsb]{202, 0.00, 1.0}{\strut adipisci}} {\setlength{\fboxsep}{0pt}\colorbox[Hsb]{350, 0.00, 1.0}{\strut tempora}} {\setlength{\fboxsep}{0pt}\colorbox[Hsb]{202, 0.01, 1.0}{\strut tracheobronchomalacia}} {\setlength{\fboxsep}{0pt}\colorbox[Hsb]{202, 0.00, 1.0}{\strut quaerat}} {\setlength{\fboxsep}{0pt}\colorbox[Hsb]{350, 0.00, 1.0}{\strut dolor.}} {\setlength{\fboxsep}{0pt}\colorbox[Hsb]{350, 0.00, 1.0}{\strut Numquam}} {\setlength{\fboxsep}{0pt}\colorbox[Hsb]{202, 0.00, 1.0}{\strut est}} {\setlength{\fboxsep}{0pt}\colorbox[Hsb]{202, 0.00, 1.0}{\strut dolore}} {\setlength{\fboxsep}{0pt}\colorbox[Hsb]{350, 0.00, 1.0}{\strut labore}} {\setlength{\fboxsep}{0pt}\colorbox[Hsb]{350, 0.00, 1.0}{\strut est}} {\setlength{\fboxsep}{0pt}\colorbox[Hsb]{202, 0.00, 1.0}{\strut neque.}} {\setlength{\fboxsep}{0pt}\colorbox[Hsb]{202, 0.09, 1.0}{\strut respiratory}} {\setlength{\fboxsep}{0pt}\colorbox[Hsb]{202, 0.06, 1.0}{\strut failure}} {\setlength{\fboxsep}{0pt}\colorbox[Hsb]{202, 0.00, 1.0}{\strut Ipsum}} {\setlength{\fboxsep}{0pt}\colorbox[Hsb]{350, 0.00, 1.0}{\strut quiquia}} {\setlength{\fboxsep}{0pt}\colorbox[Hsb]{202, 0.00, 1.0}{\strut etincidunt}} {\setlength{\fboxsep}{0pt}\colorbox[Hsb]{350, 0.00, 1.0}{\strut labore}} {\setlength{\fboxsep}{0pt}\colorbox[Hsb]{350, 0.00, 1.0}{\strut modi.}} {\setlength{\fboxsep}{0pt}\colorbox[Hsb]{350, 0.00, 1.0}{\strut Dolorem}} {\setlength{\fboxsep}{0pt}\colorbox[Hsb]{350, 0.00, 1.0}{\strut aliquam}} {\setlength{\fboxsep}{0pt}\colorbox[Hsb]{350, 0.00, 1.0}{\strut dolore}} {\setlength{\fboxsep}{0pt}\colorbox[Hsb]{350, 0.00, 1.0}{\strut amet.}} {\setlength{\fboxsep}{0pt}\colorbox[Hsb]{202, 0.00, 1.0}{\strut Amet}} {\setlength{\fboxsep}{0pt}\colorbox[Hsb]{350, 0.00, 1.0}{\strut est}} {\setlength{\fboxsep}{0pt}\colorbox[Hsb]{350, 0.00, 1.0}{\strut consectetur}} {\setlength{\fboxsep}{0pt}\colorbox[Hsb]{350, 0.00, 1.0}{\strut modi}} {\setlength{\fboxsep}{0pt}\colorbox[Hsb]{350, 0.00, 1.0}{\strut neque.}} {\setlength{\fboxsep}{0pt}\colorbox[Hsb]{350, 0.00, 1.0}{\strut Porro}} {\setlength{\fboxsep}{0pt}\colorbox[Hsb]{202, 0.01, 1.0}{\strut respiratory}} {\setlength{\fboxsep}{0pt}\colorbox[Hsb]{202, 0.03, 1.0}{\strut failure}} {\setlength{\fboxsep}{0pt}\colorbox[Hsb]{350, 0.00, 1.0}{\strut etincidunt}} {\setlength{\fboxsep}{0pt}\colorbox[Hsb]{350, 0.00, 1.0}{\strut quaerat}} {\setlength{\fboxsep}{0pt}\colorbox[Hsb]{350, 0.00, 1.0}{\strut est}} {\setlength{\fboxsep}{0pt}\colorbox[Hsb]{350, 0.00, 1.0}{\strut neque}} {\setlength{\fboxsep}{0pt}\colorbox[Hsb]{350, 0.00, 1.0}{\strut dolor}} {\setlength{\fboxsep}{0pt}\colorbox[Hsb]{350, 0.00, 1.0}{\strut quaerat.}} {\setlength{\fboxsep}{0pt}\colorbox[Hsb]{350, 0.00, 1.0}{\strut Est}} {\setlength{\fboxsep}{0pt}\colorbox[Hsb]{202, 0.00, 1.0}{\strut quaerat}} {\setlength{\fboxsep}{0pt}\colorbox[Hsb]{350, 0.00, 1.0}{\strut est}} {\setlength{\fboxsep}{0pt}\colorbox[Hsb]{350, 0.00, 1.0}{\strut adipisci}} {\setlength{\fboxsep}{0pt}\colorbox[Hsb]{350, 0.00, 1.0}{\strut ipsum.}} {\setlength{\fboxsep}{0pt}\colorbox[Hsb]{350, 0.00, 1.0}{\strut Sit}} {\setlength{\fboxsep}{0pt}\colorbox[Hsb]{350, 0.00, 1.0}{\strut dolore}} {\setlength{\fboxsep}{0pt}\colorbox[Hsb]{350, 0.00, 1.0}{\strut quisquam}} {\setlength{\fboxsep}{0pt}\colorbox[Hsb]{350, 0.00, 1.0}{\strut ipsum}} {\setlength{\fboxsep}{0pt}\colorbox[Hsb]{350, 0.00, 1.0}{\strut non}} {\setlength{\fboxsep}{0pt}\colorbox[Hsb]{350, 0.00, 1.0}{\strut neque}} {\setlength{\fboxsep}{0pt}\colorbox[Hsb]{202, 0.00, 1.0}{\strut quiquia}} {\setlength{\fboxsep}{0pt}\colorbox[Hsb]{350, 0.00, 1.0}{\strut aliquam.}} {\setlength{\fboxsep}{0pt}\colorbox[Hsb]{350, 0.00, 1.0}{\strut Ut}} {\setlength{\fboxsep}{0pt}\colorbox[Hsb]{350, 0.00, 1.0}{\strut ipsum}} {\setlength{\fboxsep}{0pt}\colorbox[Hsb]{350, 0.00, 1.0}{\strut adipisci}} {\setlength{\fboxsep}{0pt}\colorbox[Hsb]{350, 0.00, 1.0}{\strut labore}} {\setlength{\fboxsep}{0pt}\colorbox[Hsb]{350, 0.00, 1.0}{\strut tempora}} {\setlength{\fboxsep}{0pt}\colorbox[Hsb]{350, 0.00, 1.0}{\strut quaerat}} {\setlength{\fboxsep}{0pt}\colorbox[Hsb]{202, 0.00, 1.0}{\strut tempora}} {\setlength{\fboxsep}{0pt}\colorbox[Hsb]{350, 0.00, 1.0}{\strut labore.}} {\setlength{\fboxsep}{0pt}\colorbox[Hsb]{350, 0.00, 1.0}{\strut Ipsum}} {\setlength{\fboxsep}{0pt}\colorbox[Hsb]{202, 0.00, 1.0}{\strut numquam}} {\setlength{\fboxsep}{0pt}\colorbox[Hsb]{350, 0.00, 1.0}{\strut voluptatem}} {\setlength{\fboxsep}{0pt}\colorbox[Hsb]{350, 0.00, 1.0}{\strut consectetur.}} {\setlength{\fboxsep}{0pt}\colorbox[Hsb]{350, 0.00, 1.0}{\strut Aliquam}} {\setlength{\fboxsep}{0pt}\colorbox[Hsb]{350, 0.00, 1.0}{\strut voluptatem}} {\setlength{\fboxsep}{0pt}\colorbox[Hsb]{350, 0.14, 1.0}{\strut ,}} {\setlength{\fboxsep}{0pt}\colorbox[Hsb]{350, 0.00, 1.0}{\strut eius}} {\setlength{\fboxsep}{0pt}\colorbox[Hsb]{350, 0.00, 1.0}{\strut numquam.}} {\setlength{\fboxsep}{0pt}\colorbox[Hsb]{350, 0.00, 1.0}{\strut Velit}} {\setlength{\fboxsep}{0pt}\colorbox[Hsb]{202, 0.03, 1.0}{\strut generalized}} {\setlength{\fboxsep}{0pt}\colorbox[Hsb]{202, 0.00, 1.0}{\strut ut}} {\setlength{\fboxsep}{0pt}\colorbox[Hsb]{350, 0.00, 1.0}{\strut non}} {\setlength{\fboxsep}{0pt}\colorbox[Hsb]{350, 0.00, 1.0}{\strut numquam}} {\setlength{\fboxsep}{0pt}\colorbox[Hsb]{350, 0.00, 1.0}{\strut magnam}} {\setlength{\fboxsep}{0pt}\colorbox[Hsb]{350, 0.00, 1.0}{\strut sed}} {\setlength{\fboxsep}{0pt}\colorbox[Hsb]{350, 0.00, 1.0}{\strut modi.}} {\setlength{\fboxsep}{0pt}\colorbox[Hsb]{202, 0.00, 1.0}{\strut Consectetur}} {\setlength{\fboxsep}{0pt}\colorbox[Hsb]{202, 0.00, 1.0}{\strut porro}} {\setlength{\fboxsep}{0pt}\colorbox[Hsb]{350, 0.01, 1.0}{\strut .}} {\setlength{\fboxsep}{0pt}\colorbox[Hsb]{202, 0.03, 1.0}{\strut heart}} {\setlength{\fboxsep}{0pt}\colorbox[Hsb]{350, 0.00, 1.0}{\strut etincidunt}} {\setlength{\fboxsep}{0pt}\colorbox[Hsb]{202, 0.00, 1.0}{\strut eius}} {\setlength{\fboxsep}{0pt}\colorbox[Hsb]{350, 0.00, 1.0}{\strut consectetur}} {\setlength{\fboxsep}{0pt}\colorbox[Hsb]{350, 0.02, 1.0}{\strut ,}} {\setlength{\fboxsep}{0pt}\colorbox[Hsb]{350, 0.00, 1.0}{\strut quaerat}} {\setlength{\fboxsep}{0pt}\colorbox[Hsb]{350, 0.00, 1.0}{\strut amet.}} {\setlength{\fboxsep}{0pt}\colorbox[Hsb]{202, 0.00, 1.0}{\strut Amet}} {\setlength{\fboxsep}{0pt}\colorbox[Hsb]{350, 0.00, 1.0}{\strut dolorem}} {\setlength{\fboxsep}{0pt}\colorbox[Hsb]{350, 0.01, 1.0}{\strut is}} {\setlength{\fboxsep}{0pt}\colorbox[Hsb]{350, 0.19, 1.0}{\strut difficult}} {\setlength{\fboxsep}{0pt}\colorbox[Hsb]{350, 0.00, 1.0}{\strut dolor}} {\setlength{\fboxsep}{0pt}\colorbox[Hsb]{350, 0.00, 1.0}{\strut consectetur}} {\setlength{\fboxsep}{0pt}\colorbox[Hsb]{202, 0.00, 1.0}{\strut etincidunt}} {\setlength{\fboxsep}{0pt}\colorbox[Hsb]{202, 0.00, 1.0}{\strut sed}} {\setlength{\fboxsep}{0pt}\colorbox[Hsb]{202, 0.09, 1.0}{\strut effusions}} {\setlength{\fboxsep}{0pt}\colorbox[Hsb]{350, 0.00, 1.0}{\strut quiquia}} {\setlength{\fboxsep}{0pt}\colorbox[Hsb]{350, 0.00, 1.0}{\strut aliquam.}} {\setlength{\fboxsep}{0pt}\colorbox[Hsb]{350, 0.00, 1.0}{\strut Porro}} {\setlength{\fboxsep}{0pt}\colorbox[Hsb]{350, 0.00, 1.0}{\strut etincidunt}} {\setlength{\fboxsep}{0pt}\colorbox[Hsb]{350, 0.00, 1.0}{\strut dolore}} {\setlength{\fboxsep}{0pt}\colorbox[Hsb]{350, 0.00, 1.0}{\strut labore}} {\setlength{\fboxsep}{0pt}\colorbox[Hsb]{202, 0.01, 1.0}{\strut no}} {\setlength{\fboxsep}{0pt}\colorbox[Hsb]{350, 0.00, 1.0}{\strut dolore}} {\setlength{\fboxsep}{0pt}\colorbox[Hsb]{350, 0.00, 1.0}{\strut dolorem}} {\setlength{\fboxsep}{0pt}\colorbox[Hsb]{350, 0.00, 1.0}{\strut aliquam.}} {\setlength{\fboxsep}{0pt}\colorbox[Hsb]{202, 0.00, 1.0}{\strut Tempora}} {\setlength{\fboxsep}{0pt}\colorbox[Hsb]{350, 0.00, 1.0}{\strut etincidunt}} {\setlength{\fboxsep}{0pt}\colorbox[Hsb]{350, 0.00, 1.0}{\strut quisquam}} {\setlength{\fboxsep}{0pt}\colorbox[Hsb]{350, 0.00, 1.0}{\strut aliquam}} {\setlength{\fboxsep}{0pt}\colorbox[Hsb]{350, 0.00, 1.0}{\strut numquam}} {\setlength{\fboxsep}{0pt}\colorbox[Hsb]{350, 0.00, 1.0}{\strut eius}} {\setlength{\fboxsep}{0pt}\colorbox[Hsb]{202, 0.00, 1.0}{\strut ut.}} {\setlength{\fboxsep}{0pt}\colorbox[Hsb]{202, 0.13, 1.0}{\strut tracheostomy}} {\setlength{\fboxsep}{0pt}\colorbox[Hsb]{202, 0.00, 1.0}{\strut Modi}} {\setlength{\fboxsep}{0pt}\colorbox[Hsb]{350, 0.00, 1.0}{\strut modi}} {\setlength{\fboxsep}{0pt}\colorbox[Hsb]{350, 0.00, 1.0}{\strut amet}} {\setlength{\fboxsep}{0pt}\colorbox[Hsb]{350, 0.00, 1.0}{\strut voluptatem}}

\textbf{Original Output: } 0.694 \textbf{Adversarial Output: } 0.699%\vspace{-1em}
\end{minipage}     \vspace{.2em}
\hrule width\textwidth
 \vspace{.2em}
\begin{minipage}{\textwidth}
\small
\textbf{Original vs Log Odds Attention Difference : } 
{\setlength{\fboxsep}{0pt}\colorbox[Hsb]{202, 0.00, 1.0}{\strut Non}} {\setlength{\fboxsep}{0pt}\colorbox[Hsb]{350, 0.00, 1.0}{\strut magnam}} {\setlength{\fboxsep}{0pt}\colorbox[Hsb]{202, 0.00, 1.0}{\strut quiquia}} {\setlength{\fboxsep}{0pt}\colorbox[Hsb]{202, 0.00, 1.0}{\strut magnam}} {\setlength{\fboxsep}{0pt}\colorbox[Hsb]{202, 0.00, 1.0}{\strut magnam}} {\setlength{\fboxsep}{0pt}\colorbox[Hsb]{202, 0.00, 1.0}{\strut quaerat.}} {\setlength{\fboxsep}{0pt}\colorbox[Hsb]{202, 0.00, 1.0}{\strut Ut}} {\setlength{\fboxsep}{0pt}\colorbox[Hsb]{202, 0.00, 1.0}{\strut etincidunt}} {\setlength{\fboxsep}{0pt}\colorbox[Hsb]{202, 0.00, 1.0}{\strut magnam}} {\setlength{\fboxsep}{0pt}\colorbox[Hsb]{202, 0.00, 1.0}{\strut voluptatem}} {\setlength{\fboxsep}{0pt}\colorbox[Hsb]{202, 0.00, 1.0}{\strut velit}} {\setlength{\fboxsep}{0pt}\colorbox[Hsb]{202, 0.00, 1.0}{\strut eius.}} {\setlength{\fboxsep}{0pt}\colorbox[Hsb]{350, 0.00, 1.0}{\strut Dolorem}} {\setlength{\fboxsep}{0pt}\colorbox[Hsb]{202, 0.00, 1.0}{\strut dolorem}} {\setlength{\fboxsep}{0pt}\colorbox[Hsb]{202, 0.00, 1.0}{\strut velit}} {\setlength{\fboxsep}{0pt}\colorbox[Hsb]{202, 0.00, 1.0}{\strut dolor}} {\setlength{\fboxsep}{0pt}\colorbox[Hsb]{202, 0.00, 1.0}{\strut porro}} {\setlength{\fboxsep}{0pt}\colorbox[Hsb]{202, 0.00, 1.0}{\strut ut}} {\setlength{\fboxsep}{0pt}\colorbox[Hsb]{202, 0.00, 1.0}{\strut etincidunt.}} {\setlength{\fboxsep}{0pt}\colorbox[Hsb]{350, 0.00, 1.0}{\strut Consectetur}} {\setlength{\fboxsep}{0pt}\colorbox[Hsb]{202, 0.00, 1.0}{\strut dolor}} {\setlength{\fboxsep}{0pt}\colorbox[Hsb]{202, 0.00, 1.0}{\strut voluptatem}} {\setlength{\fboxsep}{0pt}\colorbox[Hsb]{202, 0.02, 1.0}{\strut cystic}} {\setlength{\fboxsep}{0pt}\colorbox[Hsb]{202, 0.04, 1.0}{\strut brain}} {\setlength{\fboxsep}{0pt}\colorbox[Hsb]{202, 0.10, 1.0}{\strut mass}} {\setlength{\fboxsep}{0pt}\colorbox[Hsb]{202, 0.00, 1.0}{\strut quaerat}} {\setlength{\fboxsep}{0pt}\colorbox[Hsb]{202, 0.02, 1.0}{\strut surgical}} {\setlength{\fboxsep}{0pt}\colorbox[Hsb]{202, 0.03, 1.0}{\strut resection}} {\setlength{\fboxsep}{0pt}\colorbox[Hsb]{202, 0.00, 1.0}{\strut est}} {\setlength{\fboxsep}{0pt}\colorbox[Hsb]{202, 0.00, 1.0}{\strut magnam}} {\setlength{\fboxsep}{0pt}\colorbox[Hsb]{350, 0.00, 1.0}{\strut etincidunt.}} {\setlength{\fboxsep}{0pt}\colorbox[Hsb]{202, 0.00, 1.0}{\strut Ipsum}} {\setlength{\fboxsep}{0pt}\colorbox[Hsb]{202, 0.00, 1.0}{\strut neque}} {\setlength{\fboxsep}{0pt}\colorbox[Hsb]{202, 0.00, 1.0}{\strut dolorem}} {\setlength{\fboxsep}{0pt}\colorbox[Hsb]{202, 0.00, 1.0}{\strut sed}} {\setlength{\fboxsep}{0pt}\colorbox[Hsb]{202, 0.00, 1.0}{\strut consectetur}} {\setlength{\fboxsep}{0pt}\colorbox[Hsb]{350, 0.00, 1.0}{\strut est.}} {\setlength{\fboxsep}{0pt}\colorbox[Hsb]{202, 0.00, 1.0}{\strut Magnam}} {\setlength{\fboxsep}{0pt}\colorbox[Hsb]{202, 0.00, 1.0}{\strut modi}} {\setlength{\fboxsep}{0pt}\colorbox[Hsb]{202, 0.00, 1.0}{\strut voluptatem}} {\setlength{\fboxsep}{0pt}\colorbox[Hsb]{350, 0.00, 1.0}{\strut dolorem}} {\setlength{\fboxsep}{0pt}\colorbox[Hsb]{202, 0.00, 1.0}{\strut tempora}} {\setlength{\fboxsep}{0pt}\colorbox[Hsb]{202, 0.00, 1.0}{\strut sed}} {\setlength{\fboxsep}{0pt}\colorbox[Hsb]{202, 0.00, 1.0}{\strut ut.}} {\setlength{\fboxsep}{0pt}\colorbox[Hsb]{202, 0.00, 1.0}{\strut Dolore}} {\setlength{\fboxsep}{0pt}\colorbox[Hsb]{350, 0.00, 1.0}{\strut dolor}} {\setlength{\fboxsep}{0pt}\colorbox[Hsb]{202, 0.00, 1.0}{\strut tempora}} {\setlength{\fboxsep}{0pt}\colorbox[Hsb]{350, 0.00, 1.0}{\strut eius}} {\setlength{\fboxsep}{0pt}\colorbox[Hsb]{202, 0.00, 1.0}{\strut aliquam}} {\setlength{\fboxsep}{0pt}\colorbox[Hsb]{350, 0.00, 1.0}{\strut quisquam.}} {\setlength{\fboxsep}{0pt}\colorbox[Hsb]{202, 0.00, 1.0}{\strut Dolor}} {\setlength{\fboxsep}{0pt}\colorbox[Hsb]{202, 0.00, 1.0}{\strut quisquam}} {\setlength{\fboxsep}{0pt}\colorbox[Hsb]{202, 0.00, 1.0}{\strut eius}} {\setlength{\fboxsep}{0pt}\colorbox[Hsb]{202, 0.00, 1.0}{\strut sed}} {\setlength{\fboxsep}{0pt}\colorbox[Hsb]{202, 0.00, 1.0}{\strut labore}} {\setlength{\fboxsep}{0pt}\colorbox[Hsb]{202, 0.00, 1.0}{\strut dolore}} {\setlength{\fboxsep}{0pt}\colorbox[Hsb]{202, 0.00, 1.0}{\strut sit}} {\setlength{\fboxsep}{0pt}\colorbox[Hsb]{350, 0.00, 1.0}{\strut velit.}} {\setlength{\fboxsep}{0pt}\colorbox[Hsb]{202, 0.00, 1.0}{\strut Magnam}} {\setlength{\fboxsep}{0pt}\colorbox[Hsb]{350, 0.00, 1.0}{\strut aliquam}} {\setlength{\fboxsep}{0pt}\colorbox[Hsb]{202, 0.00, 1.0}{\strut quisquam}} {\setlength{\fboxsep}{0pt}\colorbox[Hsb]{350, 0.00, 1.0}{\strut numquam.}} {\setlength{\fboxsep}{0pt}\colorbox[Hsb]{202, 0.00, 1.0}{\strut Aliquam}} {\setlength{\fboxsep}{0pt}\colorbox[Hsb]{202, 0.00, 1.0}{\strut sed}} {\setlength{\fboxsep}{0pt}\colorbox[Hsb]{202, 0.00, 1.0}{\strut sed}} {\setlength{\fboxsep}{0pt}\colorbox[Hsb]{202, 0.00, 1.0}{\strut modi}} {\setlength{\fboxsep}{0pt}\colorbox[Hsb]{202, 0.00, 1.0}{\strut neque.}} {\setlength{\fboxsep}{0pt}\colorbox[Hsb]{202, 0.00, 1.0}{\strut Dolor}} {\setlength{\fboxsep}{0pt}\colorbox[Hsb]{202, 0.11, 1.0}{\strut chronic}} {\setlength{\fboxsep}{0pt}\colorbox[Hsb]{202, 0.00, 1.0}{\strut quiquia}} {\setlength{\fboxsep}{0pt}\colorbox[Hsb]{202, 0.00, 1.0}{\strut voluptatem}} {\setlength{\fboxsep}{0pt}\colorbox[Hsb]{202, 0.00, 1.0}{\strut adipisci}} {\setlength{\fboxsep}{0pt}\colorbox[Hsb]{202, 0.00, 1.0}{\strut quaerat}} {\setlength{\fboxsep}{0pt}\colorbox[Hsb]{202, 0.00, 1.0}{\strut adipisci.}} \ldots\dots {\setlength{\fboxsep}{0pt}\colorbox[Hsb]{350, 0.00, 1.0}{\strut Magnam}} {\setlength{\fboxsep}{0pt}\colorbox[Hsb]{350, 0.00, 1.0}{\strut velit}} {\setlength{\fboxsep}{0pt}\colorbox[Hsb]{202, 0.00, 1.0}{\strut quaerat}} {\setlength{\fboxsep}{0pt}\colorbox[Hsb]{202, 0.00, 1.0}{\strut adipisci.}} {\setlength{\fboxsep}{0pt}\colorbox[Hsb]{202, 0.00, 1.0}{\strut Ut}} {\setlength{\fboxsep}{0pt}\colorbox[Hsb]{202, 0.01, 1.0}{\strut cystic}} {\setlength{\fboxsep}{0pt}\colorbox[Hsb]{202, 0.03, 1.0}{\strut brain}} {\setlength{\fboxsep}{0pt}\colorbox[Hsb]{202, 0.08, 1.0}{\strut mass}} {\setlength{\fboxsep}{0pt}\colorbox[Hsb]{202, 0.00, 1.0}{\strut adipisci}} {\setlength{\fboxsep}{0pt}\colorbox[Hsb]{202, 0.00, 1.0}{\strut velit}} {\setlength{\fboxsep}{0pt}\colorbox[Hsb]{202, 0.00, 1.0}{\strut modi.}} {\setlength{\fboxsep}{0pt}\colorbox[Hsb]{202, 0.00, 1.0}{\strut Sed}} {\setlength{\fboxsep}{0pt}\colorbox[Hsb]{202, 0.00, 1.0}{\strut aliquam}} {\setlength{\fboxsep}{0pt}\colorbox[Hsb]{202, 0.02, 1.0}{\strut astrocytoma}} {\setlength{\fboxsep}{0pt}\colorbox[Hsb]{202, 0.00, 1.0}{\strut est}} {\setlength{\fboxsep}{0pt}\colorbox[Hsb]{202, 0.00, 1.0}{\strut porro.}} {\setlength{\fboxsep}{0pt}\colorbox[Hsb]{202, 0.00, 1.0}{\strut Labore}} {\setlength{\fboxsep}{0pt}\colorbox[Hsb]{202, 0.01, 1.0}{\strut resection}} {\setlength{\fboxsep}{0pt}\colorbox[Hsb]{202, 0.00, 1.0}{\strut eius}} {\setlength{\fboxsep}{0pt}\colorbox[Hsb]{202, 0.00, 1.0}{\strut voluptatem}} {\setlength{\fboxsep}{0pt}\colorbox[Hsb]{202, 0.00, 1.0}{\strut sit}} {\setlength{\fboxsep}{0pt}\colorbox[Hsb]{350, 0.00, 1.0}{\strut quisquam}} {\setlength{\fboxsep}{0pt}\colorbox[Hsb]{202, 0.00, 1.0}{\strut consectetur}} {\setlength{\fboxsep}{0pt}\colorbox[Hsb]{202, 0.00, 1.0}{\strut modi.}} {\setlength{\fboxsep}{0pt}\colorbox[Hsb]{202, 0.00, 1.0}{\strut Est}} {\setlength{\fboxsep}{0pt}\colorbox[Hsb]{202, 0.00, 1.0}{\strut ipsum}} {\setlength{\fboxsep}{0pt}\colorbox[Hsb]{350, 0.48, 1.0}{\strut tumor}} {\setlength{\fboxsep}{0pt}\colorbox[Hsb]{202, 0.00, 1.0}{\strut dolore}}

\textbf{Original Output: } 0.798 \textbf{Log Odds Output : } 0.800\vspace{-.5em}
%sarthak: I have removed the caption for adversarial case separately. Lets wrap up the caption for both in the same.
\caption{Heatmaps showing difference in Original and counterfactual attention distributions over clinical notes from MIMIC, where we have replaced text with \emph{lorem ipsum} for all but the most relevant tokens in order to preserve privacy (red implies counterfactual attention is higher and blue vice-versa). These show different cases where we can significantly change the attention distribution (either \textbf{adversarial (Top)} or using \textbf{Log Odds (Bottom)} while barely affecting the prediction.}\vspace{-.5em}
\end{minipage}
\end{figure*}

\EvalTabs

\subsection{Counterfactual Experiments}
We investigate if model predictions \emph{would} have differed, had the model attended to different words (i.e., under \emph{counterfactual} attention distributions).  

\begin{figure*}[!h]
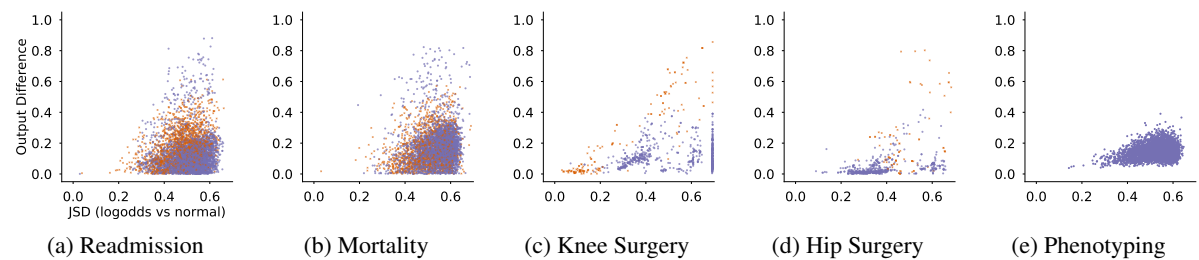

    \centering
    \AddSubFigureForAll{graph_outputs/logodds_subs_Scatter}
    \vspace{-.5em}
    \caption{Change in output ($y$-axis) using original attention vs Log Odds attention during predictions against JSD between these two distributions ($x$-axis). These results are for LSTM encoders.}
    \label{fig:logoddssub_out}
    \vspace{-.5em}
\end{figure*}

We follow the two strategies from \cite{jain2019attention} for constructing counterfactual attention distributions. In the first we randomly permute the empirical weights obtained from the attention module prior to inducing the weighted representation $\vect{h}_\alpha$. We repeat this process 100 times and record the median change in output. 

The second strategy is \emph{adversarial}; we explicitly aim to identify attention weights that are maximally different from the observed weights, with the constraint that this does not change the model output by more some small value $\epsilon$. In both cases, all other model parameters are held constant.

In Figures~\ref{fig:PermutationScatter} and \ref{fig:AdversarialScatter}, we observe that predictions are unchanged under alternative attention configurations in a significant majority of cases across all architectures. Thus, attention cannot be viewed casually in the sense of `the model made these predictions \emph{because} these words were attended to'. Alternative attention distributions that yield equivalent predictions would seem to be equally plausible under the view of attention as explanation.

\subsection{Log Odds Experiments}
As a novel exercise, we also consider swapping log-odds scores for features (from an LR model operating over BoW) in for attention weights in BiLSTM model. Specifically, we induce a `log odds attention' over an input by substituting the absolute value of log odds (as estimated via LR) of the word present at each position and passing this through a softmax: $\vect{\alpha}^{LO} = \text{softmax}_t(\{\beta_{w_t}\}_{t=1}^T)$ where $w_t$ is the word at position $t$ and $\beta$ are log-odds estimates. 

These scores enjoy a clear interpretation under a linear regime. We thus explore two ways of using them with attentive neural models: (1) Swapping in these in as attention weights place of $\vect{h}_\alpha$ at test (prediction) time; (2) Use the (fixed) `log-odds attention' during training, in place of learning the attention distribution end-to-end.  

Table~\ref{tab:evals} shows that using log odds attention at test time does not degrade the performance significantly in most datasets (and actually improves performance for the Knee Surgery Complications task). Similarly, using log odds attention during training also yields similar performance to standard attention variants. But as we see in Figure~\ref{fig:logoddssub_out}, log odds attention distributions can differ considerably from learned attention distributions, again highlighting the difficulty of interpreting attention weights. 

\section{Discussion and Conclusions}

Across two EMR datasets and five predictive tasks, we have shown that (i) attention mechanisms substantially boost the performance of LSTM text encoders passed over clinical notes, but, (ii) treating attention weights as `explanations' for predictions is unwarranted. The latter confirms that the recent general findings of \cite{jain2019attention} hold in the clinical domain; this is important because interpretability in this space is critical for obvious reasons. 

We hope that this paper inspires work on transparent attention mechanisms for models that make predictions on the basis of EMR.

\section*{Acknowledgments}

This work was supported by the Army Research Office (ARO), award W911NF1810328. 

\bibliography{naaclhlt2019}
\bibliographystyle{acl_natbib}

\clearpage 
\newpage
\onecolumn
\begin{center}
{\Large \textbf{An Analysis of Attention over Clinical Notes for Predictive Tasks: Appendix}}
\end{center}
\appendix
\newcommand{\loadtex}[2]{
\clearpage
\section{#2}
\subsection{Adversarial Example}
\input{graph_outputs/adv_examples_pos/#1+lstm+tanh.csv}
\subsection{Log Odds Attention (Maximally different)}
\input{graph_outputs/adv_examples_odds_pos/#1+lstm+tanh.csv}
% \subsection{Log Odds Attention (Maximally similar)}
% \input{graph_outputs/adv_examples_odds_pos_rev/#1+lstm+tanh.csv}
}

% \onecolumn
\section{Dataset Statistics}
\begin{table*}[!htbp]
\centering
    \begin{tabular}{l c c || c c }
         \emph{Task} & $\boldsymbol{|V|}$ & \emph{Avg. length} &
         \emph{Train size} & \emph{Test size}  \\
          \hline
         Readmission & 36464 &  3865 & 23790 / 5499 & 4265 / 735  \\
        Mortality & 34030 & 3901 & 21347 / 4675 & 4323 / 677  \\
        Hip Surgery Complications & 10842 & 2624 & 3281 / 369 & 719 / 75  \\
        Knee Surgery Complications & 10842 & 2586 & 2664 / 324 & 582 / 48 \\
        Phenotyping & 10842 & 3641 & 31075 & 5000  
    \end{tabular}
    \caption{Dataset characteristics. For train and test size, we list the cardinality for each class, where applicable: $0$/$1$ for binary classification and overall for multilabel. Average length is in tokens.} 
    \label{tab:datastats}
    \vspace{-1em}
\end{table*}

The Phenotypes studied in Phenotyping task are - 

 Acute and unspecified renal failure,
 Acute cerebrovascular disease,
 Acute myocardial infarction,
 Cardiac dysrhythmias,
 Chronic kidney disease,
 Chronic obstructive pulmonary disease and bronchiectasis,
 Complications of surgical procedures or medical care,
 Conduction disorders,
 Congestive heart failure - nonhypertensive,
 Coronary atherosclerosis and other heart disease,
 Diabetes mellitus with complications,
 Diabetes mellitus without complication,
 Disorders of lipid metabolism,
 Essential hypertension,
 Fluid and electrolyte disorders,
 Gastrointestinal hemorrhage,
 Hypertension with complications and secondary hypertension,
 Other liver diseases,
 Other lower respiratory disease,
 Other upper respiratory disease,
 Pleurisy - pneumothorax - pulmonary collapse,
 Pneumonia (except that caused by tuberculosis or sexually transmitted disease),
 Respiratory failure - insufficiency - arrest (adult),
 Septicemia (except in labor),
 Shock .

\section{Model Details}
For all datasets, we use \href{https://spacy.io/}{spaCy} for tokenization. We map out of vocabulary words to a special $\texttt{<unk>}$ token and map any word with numeric characters to `qqq'. Each word in the vocabulary was initialized using pretrained embeddings \cite{moen2013distributional}. We initialize words not present in the vocabulary using samples from a standard Gaussian ($\mu=0$, $\sigma^2=1$).

\subsection{BiLSTM}
We use an embedding size of 300 and hidden size of 128 for all datasets. The model was regularized with $L_2$ regularization ($\lambda = 10^{-5}$) applied to all parameters. We use a sigmoid activation function for all binary classification tasks. We treat each phenotype classification as binary classification and take the mean loss over labels during training. We trained the model using maximum likelihood loss function with Adam Optimizer with default parameters in PyTorch.

\subsection{CNN}
We use an embedding size of 300 and 4 kernels of sizes [1, 3, 5, 7], each with 64 filters, giving a final hidden size of 256. We use ReLU activation function on the output of the filters. All other configurations remain same as BiLSTM.

\subsection{Average}
We use the embedding size of 300 and a projection size of 256 with ReLU activation on the output of the projection matrix. All other configurations remain same as BiLSTM.

% \loadtex{readmission}{Readmission}
% \loadtex{mortality}{Mortality}
% \loadtex{KneeSurgery}
% \loadtex{HipSurgery}
% \loadtex{Phenotyping}{Phenotyping}

\end{document}